\documentclass[conference]{IEEEtran}
\usepackage{cite}
\ifCLASSINFOpdf
   \usepackage[pdftex]{graphicx}
\else
\fi
\usepackage{amsmath}
\usepackage{algorithmic}
\usepackage{esvect}

\usepackage{array}
\usepackage{multirow}

\usepackage{caption}
\usepackage{cite}

\usepackage{fancyhdr}

\begin{document}
%\IEEEoverridecommandlockouts
%\IEEEpubid{\makebox[\columnwidth]{978-1-4799-2442-4/13/\$31.00 \copyright 2013 IEEE \hfill} \hspace{\columnsep}\makebox[\columnwidth]{ }}

\title{FPGA-based ORB Feature Extraction for Real-Time Visual SLAM}

\author{\IEEEauthorblockN{Weikang Fang\IEEEauthorrefmark{1}\IEEEauthorrefmark{2},
Yanjun Zhang\IEEEauthorrefmark{1},
Bo Yu\IEEEauthorrefmark{2},
Shaoshan Liu\IEEEauthorrefmark{2},
\IEEEauthorblockA{\IEEEauthorrefmark{1} Institute of Application Specific Instruction Set Processor,\\
School of Electronics and Information, Beijing Institute of Technology, Beijing, 100081, China}
\IEEEauthorblockA{\IEEEauthorrefmark{2} PerceptIn, Shenzhen, 518046, China}
}}
\maketitle

\thispagestyle{fancy}
\fancyhead{}
\lhead{}
\lfoot{\textbf{\emph{978-1-5090-4825-0/17/\$31.00~\copyright2017 IEEE}}}
\cfoot{}
\rfoot{}

\begin{abstract}
Simultaneous Localization And Mapping (SLAM) is the problem of constructing or updating a map of an unknown environment while simultaneously keeping track of an agent's location within it. How to enable SLAM robustly and durably on mobile, or even IoT grade devices, is the main challenge faced by the industry today. The main problems we need to address are: 1.) how to accelerate the SLAM pipeline to meet real-time requirements; and 2.) how to reduce SLAM energy consumption to extend battery life. After delving into the problem, we found out that feature extraction is indeed the bottleneck of performance and energy consumption. Hence, in this paper, we design, implement, and evaluate a hardware ORB feature extractor %, and demonstrate that our design not only outperform existing SoCs including ARM Krait and Intel Core i5. More importantly, we
and prove that our design is a great balance between performance and energy consumption compared with ARM Krait and Intel Core i5.   %such that we can tune the parameters in the design to achieve greater performance, or energy efficiency.
\end{abstract}

\begin{IEEEkeywords}
ORB, feature extraction, SLAM, FPGA
\end{IEEEkeywords}

% \fi
\IEEEpeerreviewmaketitle

\section{Introduction}
\begin{figure}[!b]
\setlength{\belowcaptionskip}{-0.00in}
\centering
\includegraphics[width=2.8in]{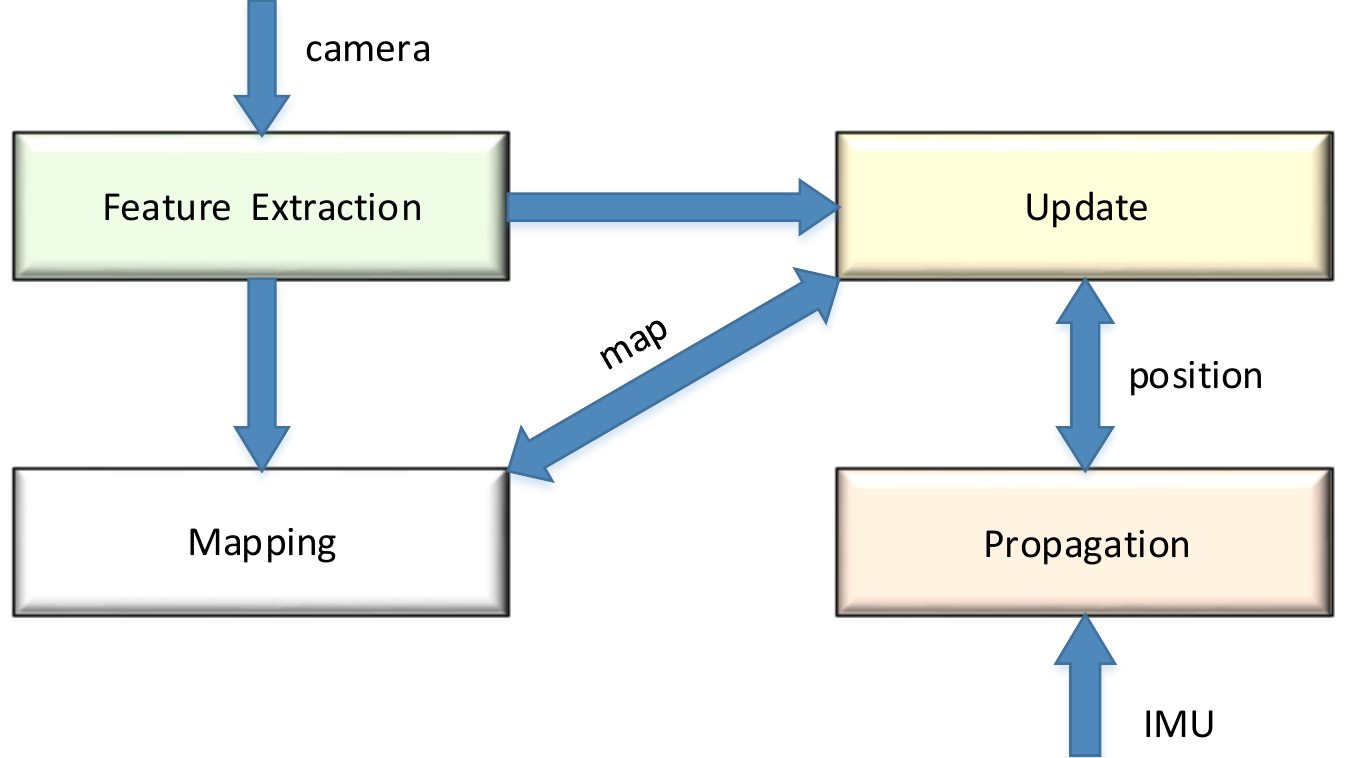}
\caption{{Overview of a visual inertial SLAM system.}}
\label{fig:salm}
\end{figure}

Simultaneous Localization And Mapping (SLAM) [7]-[9] is the core enabling technology behind applications such as autonomous vehicles, robotics, virtual reality (VR), and augmented reality (AR). %SLAM technology imposes two main challenges: first, SLAM has a complex pipeline and is computation intensive, and often it imposes strong real-time requirements. %For instance, in the case of Visual SLAM, camera image data can rush in at a rate as high as 60 FPS, meaning that the computation pipeline needs to process each image within 16 ms in order for the system not to lose track of itself. In addition, the images samples form a time series such that they cannot be processed in parallel.
%Second, these workloads mostly run on battery-powered mobile devices with extremely limited energy budget. %Many VR and AR applications run on mobile phones today and drain battery quite rapidly.  %For example, our experiments showed that the Google Tango device, when running AR applications, would see its battery drain within forty minutes.
In detail, SLAM is the problem of constructing or updating a map of an unknown environment while simultaneously keeping track of an agent's location within it. Fig. \ref{fig:salm} shows a simplified version of our production-level Visual Inertial SLAM system which utilizes design concepts from MSCKF [1] and ORB-SLAM [2]. %we have shipped several generations of robotic products based on this SLAM technology.

We implemented this SLAM technology on a quad-core ARM v8 mobile SoC. Based on our profiling results, the feature extraction stage is indeed the most computation-intensive, consuming ${>}50\%$ of the CPU resources.  Even with such heavy computation resource consumption, it still takes about 50 ms to extract features from a VGA-resolution ($640\times480$) image, leading to a frame rate only at about 20 FPS. %This brings three problems: first, we would like the frame rate to be higher to make the robotic system more robust and less prone to losing track of itself; second, to make the matter more severe, we may use higher resolution images in the near future, this could lead to further reduction of frame rate;  third, feature extraction alone takes a significant portion of the overall system energy budget and affects the system’s durability.

Therefore, in this paper, we aim to address these practical problems down to the hardware level, by implementing a hardware accelerator of ORB feature extractor. ORB is widely used in robotics and it is proven to provide a fast and efficient alternative to SIFT [3]. %and have it fit in our real-time SLAM system.
In addition, we examine its performance, resource occupation, and energy consumption of the design in order to achieve further improvements.

The rest of this paper is organized as follows. In Section \uppercase\expandafter{\romannumeral2}, we review the background information of ORB algorithm. In Section \uppercase\expandafter{\romannumeral3} we describe the architecture of the proposed hardware feature extractor. In Section \uppercase\expandafter{\romannumeral4}, we share the experimental methodologies and results. Finally, we summarize our conclusions in Section \uppercase\expandafter{\romannumeral5}.

\begin{figure}[!t]
\setlength{\belowcaptionskip}{-0.20in}
\centering
\includegraphics[width=3.5in]{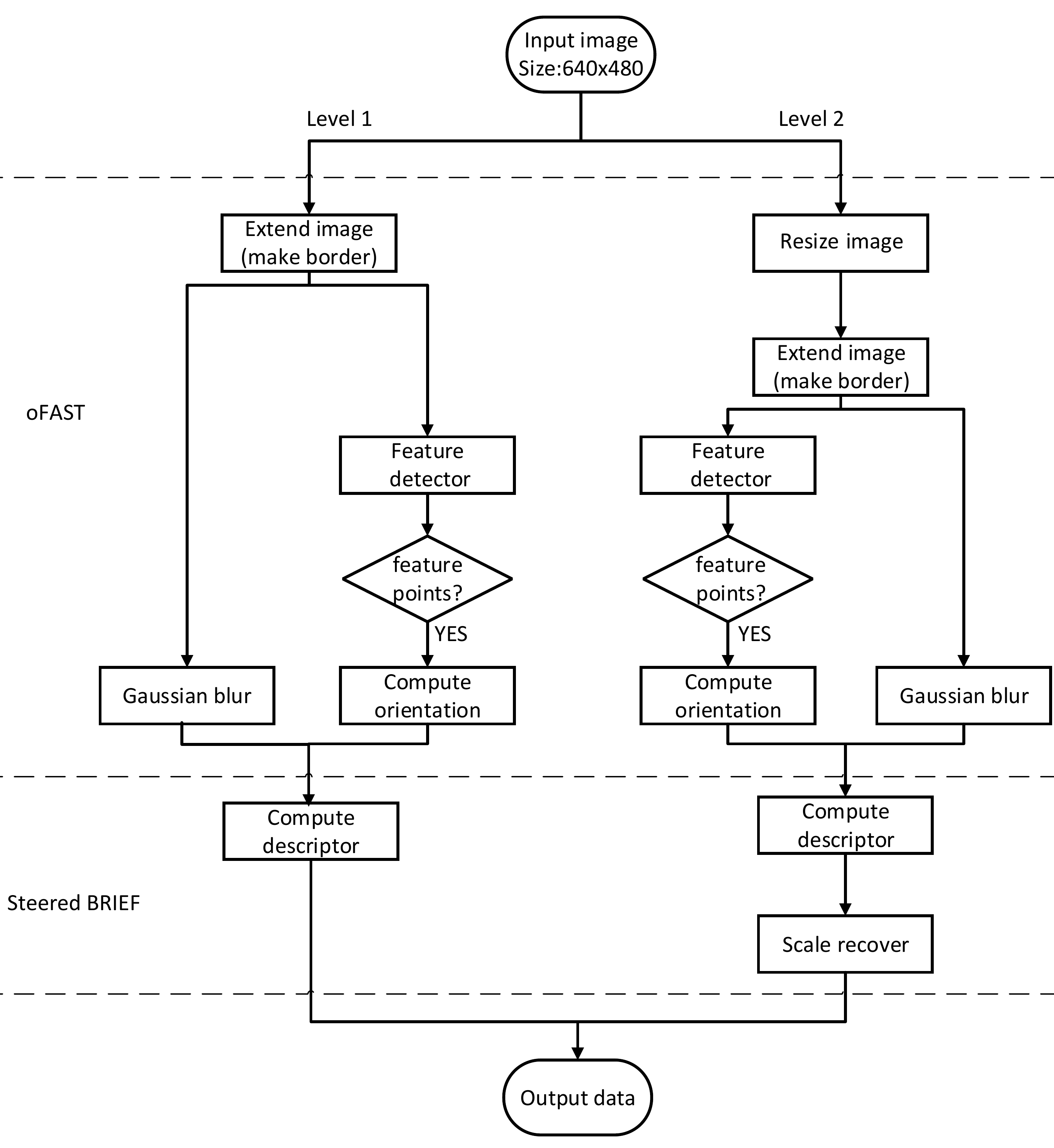}
\caption{ORB based feature extraction algorithm.}
\label{fig:orb}
\end{figure}


\begin{figure*}[!t]
\setlength{\belowcaptionskip}{-0.10in}
\centering
\includegraphics[width=6.5in]{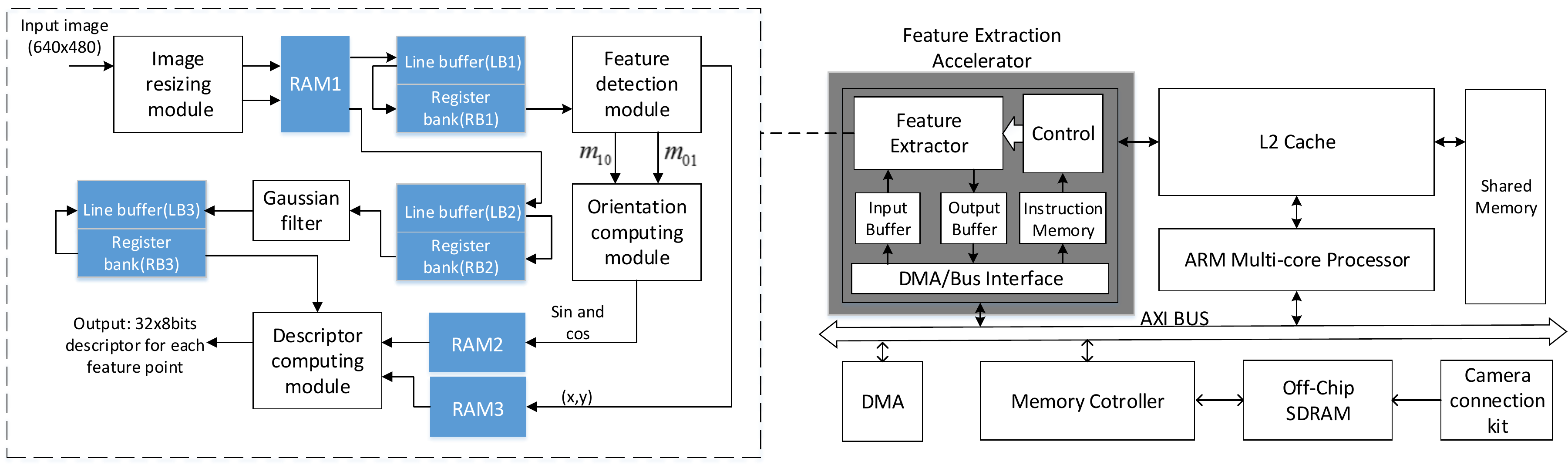}
\caption{Hardware architecture of real-time SLAM system (right) and hardware architecture of ORB feature extraction accelerator (left).}
\label{fig:fpga_arch}
\end{figure*}

\section{ORB Feature Extraction Algorithm for Visual SLAM}
The overview of ORB based feature extraction algorithm is shown in Fig. \ref{fig:orb}. %The ORB feature extraction algorithm
It consists of two parts, oFAST (Oriented Feature from Accelerated Segment Test)
[5] based feature detection and BRIEF (Binary Robust Independent Elementary Features)
based feature descriptors computation [6]. %Details of oFAST feature detector and steered BRIEF feature descriptor are described in the following sections.

%The overview of ORB based feature extraction algorithm is shown in Fig. \ref{fig:orb}. The input of the algorithm is a gray scale image with the size of $640\times480$ pixels. An $n$-level image pyramid is computed based on the input image to make the algorithm scale invariant[4]. %To obtain a $n$-level pyramid, each level of image is obtained by scaling the previous level of image down by a particular factor.
%The border of resized images is extended to ensure the correctness of feature detection. The number of levels of the pyramid, $n$, is a parameter in ORB algorithm. In our design, $n$ is 2, which enables a economic hardware implementation and does not affect the performance of the algorithm.
%After building the image pyramid, a FAST feature detector is employed to compute feature points of images.  After feature detection, the orientation of each feature point is calculated to make the algorithm rotation invariant. Each level of image is smoothed by the Gaussian filter to improve accuracy. Steered BRIEF is used to compute feature descriptors for each level of image. Lastly, feature descriptors of resized images are scaled back to the size of input image. The output of ORB algorithm is a serial of feature points and descriptors. Details of oFAST feature detector and steered BRIEF feature descriptor are described in the following sections.

\subsection{Oriented Feature from Accelerated Segment Test (oFAST)}
In general, given %a point in
an image, oFAST helps find out feature points and then calculates orientations of feature points to ensure rotation invariant. % calculates intensity differences between the point and other points around it. Points that differ greatly to the reference point in intensity are feature points. Besides coordinates of feature points, orientations of feature points are calculated to ensure rotation invariant.
Details of oFAST are shown in Fig. 2. %illustrated as follows.
First, resize the original image %compute an $n$-level image pyramid
using bilinear interpolation [4]. %In Fig. \ref{fig:orb}, the input image is resized only once to build a 2-level image pyramid.
Second, compute feature points of both original image and resized image. %each level of image by feature detector.
Third, determine the orientation of feature points. The orientation of feature points are computed as follows.

Define the patch of a feature point be the circle centered at the feature point. The moments of the patch, $m_{pq}$, are defined as,
\begin{equation}
m_{pq} = \sum_{x,y\in{r}}^{} x^p y^q I(x,y) \qquad p,q = 0\;or\;1
\end{equation}
where $I(x,y)$ is the intensity value of the point $(x,y)$ in the patch and $r$ is the radius of the circle. %The radius of the patch is 15 pixels in our design. Let the feature point be the original point, the coordinate of the intensity centroid of the patch is defined as, $(\frac{m_{10}}{m_{00}},\frac{m_{01}}{m_{00}})$.
The orientation of the feature point, $\theta$, is obtained by, $\theta = \arctan(\frac{m_{01}}{m_{10}})$. $\sin\theta$ and $\cos\theta$ is calculated as follows. %which is used to compose rotation matrices.

\begin{equation}
\sin\theta = \frac{m_{01}}{\sqrt{m_{10}^2 + m_{01}^2}} \quad \cos\theta = \frac{m_{10}}{\sqrt{m_{10}^2 + m_{01}^2}}
\label{eq:sin}
\end{equation}

%\begin{equation}
%\theta = \arctan(\frac{m_{01}}{m_{10}})
%\label{eq:rotation}
%\end{equation}

\subsection{Steered Binary Robust Independent Elementary Features (steered BRIEF)}
In general, a feature point is represented by a set of descriptor. In ORB algorithm, steered BRIEF algorithm
is employed to compute descriptors of feature points. The details of steered BRIEF algorithm
is described as follows.
%\begin{figure}[!h]
%\setlength{\belowcaptionskip}{-0.05in}
%\centering
%\includegraphics[width=2in]{CD2.pdf}
%\caption{Compute descriptors of feature point P.}
%\label{fig:steered}
%\end{figure}

Firstly, consider the circular patch defined in oFAST. %that has the radius of $r$ and is centered at the feature point P.
Select $M$ pairs of points in the patch according to Gaussian distribution. Secondly, in order to ensure rotation invariant, rotate these pairs of points by the angle determined by equation 2. %which is illustrated in Fig. \ref{fig:steered}. For a clear illustration, only four pairs of points are shown in Fig. \ref{fig:steered}. %In our design, 256 pairs of points are selected from the patch.
%In the figure,
%For example, when $M=4$, four pairs of points
Thus, after rotation, $M$ pairs of points could be labeled as $D_1(I_{A_1},I_{B_1})$,$D_2(I_{A_2},I_{B_2})$,$D_3(I_{A_3},I_{B_3})$ ... $D_M(I_{A_M},I_{B_M})$, where $A_i$ and $B_i$ are the two points of a pair, $I_{A_i}$ and $I_{B_i}$ are intensity values of the point. Thirdly, an operator is defined as follows,
\begin{equation}
T(D_i(I_{A_i},I_{B_i})) = \left\{
                      \begin{aligned}
                       1 & \qquad if I_{A_i}\geq I_{B_i} \\
                       0 & \qquad if I_{A_i} < I_{B_i}
                      \end{aligned}
\right. ,
\end{equation}
For each pair of points, the operator, $T$, produces one bit of result. %In our illustration,
With $M$ pairs of points, %When $M=4$,
the operator, $T$, produces a bit vector with the length of $M$. For example, $T$ produces the following results, $T(D_1(I_{A_1},I_{B_1}))=1$, $T(D_2(I_{A_2},I_{B_2}))=1$, $T(D_3(I_{A_3},I_{B_3}))=0$ ... $T(D_M(I_{A_M},I_{B_M}))=1$, then the descriptor of the feature point P is $110...1$. The bit vector is descriptor of the feature points.

%\begin{displaymath}
%\begin{aligned}
%T(D_1(I_{A_1},I_{B_1}))=1 \\
%T(D_2(I_{A_2},I_{B_2}))=1 \\
%T(D_3(I_{A_3},I_{B_3}))=0 \\
%T(D_4(I_{A_4},I_{B_4}))=1 \\
%\end{aligned}
%\end{displaymath}

%In order to ensure rotation invariant, pixels in patch centered around the feature point should be rotated by the %a particular
%angle determined by equation \ref{eq:sin}. Since rotating all the points in the patch is computationally expensive, our design only rotates the pairs of points that are used for computing descriptors. %Suppose that $n$ pairs of points are selected to generate descriptors, a 2$\times$n matrix consisting of these points is defined as,
%\begin{equation}
%\vec{S} = \left[
%\begin{matrix}
%x_1,x_2,\ldots,x_{2n-1},x_{2n}\\
%y_1,y_2,\ldots,y_{2n-1},y_{2n}
%\end{matrix}
%\right]
%\end{equation}
%The rotation matrix is defined as,
%\begin{equation}
%\vec{R_\theta} = \left[
%\begin{matrix}
%\quad \cos\theta \quad \sin\theta & \quad \\
%-\sin\theta \quad \cos\theta & \quad
%\end{matrix}
%\right].
%\end{equation}
%$\sin\theta$ and $\cos\theta$ are determined by equation \ref{eq:sin}. The rotated results are obtained by,
%The rotated version of $\vec{S}$ matrix is
%\begin{equation}
%\vec{S_\theta} = \vec{R_\theta}\vec{S}
%\end{equation}
%where $\vec{S_\theta}$ is the result of rotation. After rotation, the points in $\vec{S_\theta}$ are used to obtain the feature descriptors.

\section{Hardware Architecture}

\subsection{Overall Architecture of FPGA-based Real-Time SLAM}
Fig. \ref{fig:fpga_arch} (right) illustrates the proposed FPGA-based architecture of real-time SLAM system. %ORB feature extraction is implemented by the reconfigurable logic, which is able to fully explore the parallelism in the ORB algorithm and improve the performance of feature extraction.
ARM Multi-core processors are used to perform control logic and other computations in SLAM. AXI bus is employed to connected memory systems, ORB accelerator and ARM processors.
%\begin{figure*}[!t]
%\setlength{\belowcaptionskip}{-0.02in}
%\centering
%\includegraphics[width=6.5in]{1122.pdf}
%\caption{Hardware architecture of real-time SLAM system (right) and hardware architecture of ORB feature extraction %accelerator (left).}
%\label{fig:fpga_arch}
%\end{figure*}

The feature extraction accelerator consists of a data accessing part and a kernel part. The kernel part is composed of Instruction Memory, Control unit and Feature Extractor. The data accessing part is composed of Input Buffer, Output Buffer and DMA/Bus Interface. The DMA/Bus Interface directly accesses data stream captured by the camera via AXI Bus and store data in the Input Buffer for the Feature Extractor. The results of the Feature Extractor are stored in the Output Buffer and are finally sent to L2 cache. The ARM multi-core System would use these features for further processing.% such as, mapping, updating and propagating.

\subsection{Hardware Architecture of ORB Feature Extractor}
Hardware architecture of ORB feature extractor is depicted in Fig. \ref{fig:fpga_arch} (left). The architecture consists of the image resizing module, the feature detection module, the orientation computing module, and the descriptor computing module. Line buffers and registers are used to store intermediate results.
\begin{figure*}[!t]
\setlength{\belowcaptionskip}{-0.10in}
\centering
\includegraphics[width=6.6in]{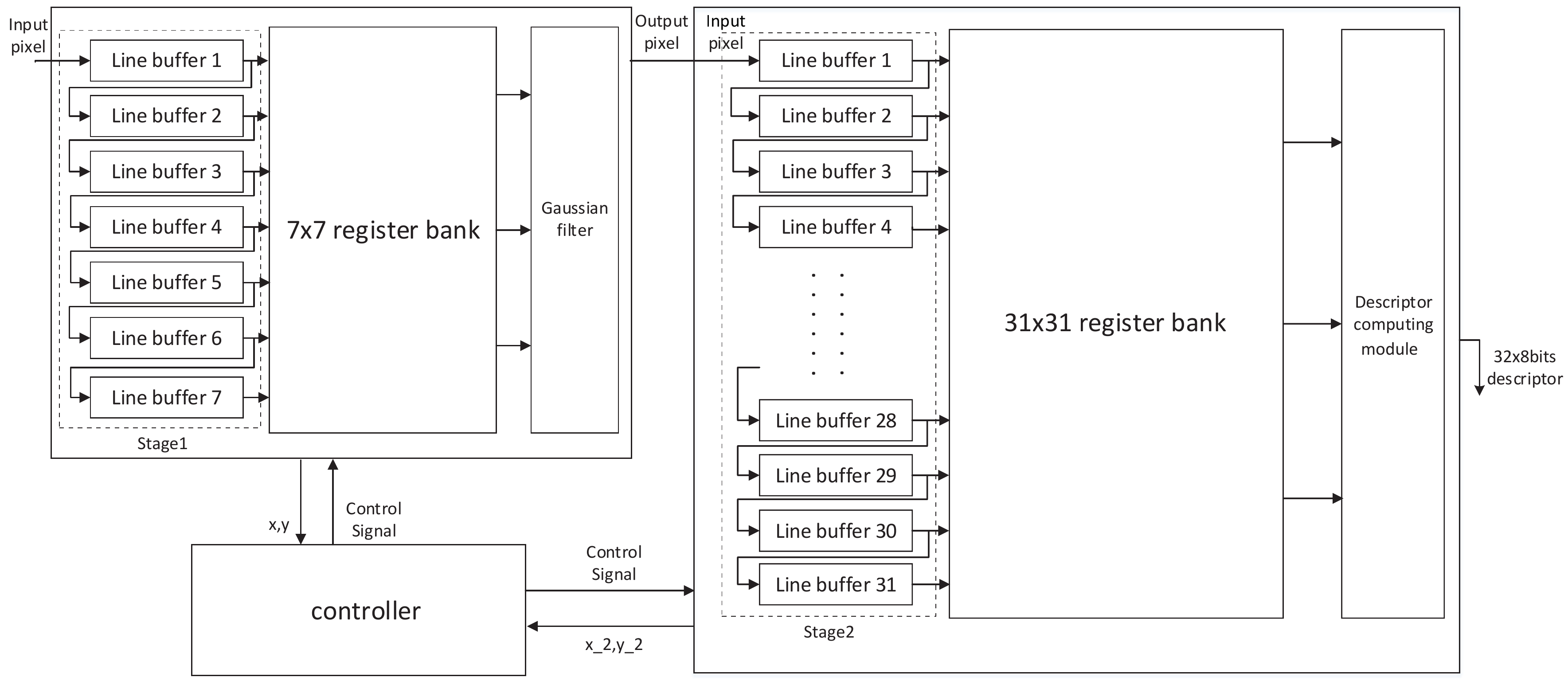}
\caption{Hardware architecture of the synchronized two-stages line shifting buffers}
\label{fig:hardware_des}
\end{figure*}

%With the scale factor SF=1.2,
The resizing module implements bilinear interpolation and builds a $2$-level image pyramid. The size of an input image is $640\times480$ and the size of a scaled image is $533\times400$. RAM1 stores the image of each level.

The LB1 (line buffer 1) stores 31 lines of pixels. RB1 (register bank 1) stores $31\times31$ pixels of neighborhood that contains a circular patch for each point. The feature detection module detects the feature point and computes %the intensity centroid
the moments of a patch, i.e. $m_{10}$ and $m_{01}$. The coordinates $(x,y)$ of the feature points are stored in RAM3.
%Fig. \ref{fig:hardware_orientation} depicts the hardware architecture of the orientation computing module. It consists of two multipliers, two fixed point dividers and a fix point square root unit. The input of orientation computing module is the intensity centroid of a patch. It

The orientation computing module computes $\sin\theta$ and $\cos\theta$ according to equation \ref{eq:sin}. $\sin\theta$ and $\cos\theta$ are stored in RAM2. %Implementing division and square root hardware requires substantial hardware costs. A word length optimization method is utilized to reduce the hardware consumption and evaluation results are shown in section \uppercase\expandafter{\romannumeral4}.
%\begin{figure}[!t]
%\setlength{\belowcaptionskip}{-0.17in}
%\centering
%\includegraphics[width=3in]{3414.pdf}
%\caption{Hardware architecture of orientation computing module.}
%\label{fig:hardware_orientation}
%\end{figure}
A $7\times7$ hardware Gaussian filter is used to smooth images. LB2 and RB2 buffers 7 lines of pixels and a $7\times7$ patch of pixels for the Gaussian filter respectively. LB3 and RB3 are used to store results of Gaussian filter. Since the Descriptor computing module calculates descriptors in a $31\times31$ patch of pixels, LB3 stores 31 lines of pixels and RB3 buffers $31\times31$ bytes.

The descriptor computing module takes smoothed images, orientations and coordinates of feature points in the image as inputs and calculates descriptors of feature points. Since the descriptor computing module depends on the results of the Gaussian filter, a straightforward implementation is to buffer the whole image before starting to compute descriptors. However, this straightforward design that needs to store the whole image requires substantial on-chip memory resources. Furthermore, starting to compute descriptors after Gaussian filtering on the whole image will stall the stream processing of the feature detection and orientation computation, which will increase the latency of the system.

The synchronized two-stages shifting line buffers, as shown in Fig. \ref{fig:hardware_des}, is proposed to compute Gaussian filtering and descriptors in a streaming way. The first stage and the second stage of line buffers store pixels for the Gaussian filter and the descriptor computing module respectively. Line buffers in each stage are organized as a shifter. The control unit synchronizes the data movement in the line buffers. If no feature point is detected, the two-stages line buffers shifts data in the buffers. If a feature point is detected, the control unit stops shifting the two-stages line buffers and starts to compute the descriptor for the feature point. By utilizing the proposed synchronized two-stages line buffers, 575K bytes of on-chip memory are saved.

%\begin{figure*}[!t]
%\setlength{\belowcaptionskip}{-0.10in}
%\centering
%\includegraphics[width=6in]{2121.pdf}
%\caption{Hardware architecture of the synchronized two-stages line shifting buffers}
%\label{fig:hardware_des}
%\end{figure*}

\section{Evaluation Results and Discussion}

\subsection{Word length optimization}
According to equation \ref{eq:sin}, the orientation of a feature point is determined by the moments of a circular patch. In our design, the radius of a patch is $15$. The range of the value of $m_{10}$ and $m_{01}$ is from $-624750$ to $624750$. 21 bits are needed to represent $m_{10}$ and $m_{01}$. The %multiplier, divider and square root hardware
orientation computing module requires even more bits to keep precision, which will consume substantial hardware resources.

Word length optimization is used to reduce hardware consumption. By computing $m_{10}$ and $m_{01}$ of all possible circular patches, we find that many significant bits in $m_{10}$ and $m_{01}$ are zeros in most cases. Besides, the influence of low bits in the calculation is quite small. Therefore, the proposed operation to shorten the word length is described as follow.

First, starting from the highest data bit without considering sign bit, find the overlapping 0s of $m_{10}$ and $m_{01}$ then remove them. Second, in the remaining data bits, take the higher N bits. If remaining part is less than N bits, fill it with zeros. Finally, splice the sign bit and obtained N bits together. In this way, the word length would be shorten to N+1 bits.

Truncating the word length introduces truncation errors in computing rotation angles and further affects the coordinates of points after rotating. For a particular point $(x, y)$, the coordinates after rotating is $(x',y')$ that is determined by,

\begin{equation}
\begin{aligned}
x' = x\cdot\cos\theta + y\cdot\sin\theta, \\
y' = y\cdot\cos\theta - x\cdot\sin\theta.
\end{aligned}
\end{equation}
We define the following metric to quantitatively evaluate the error introduced by word length truncation,

\begin{equation}
Error = \sqrt{(x' - x'_N)^2 + (y' - y'_N)^2}
\label{eq:error}
\end{equation}
where $(x',y')$ is the coordinate calculated with original word length while $(x_{N}',y_{N}')$ is calculated with the word length of $N$. %This error represents the change in the patch introduced by the word length truncation.

\begin{table}[!b]
\caption{Hardware comsumption of the Orientation computing unit}
\label{table:word_len}
\centering
\begin{tabular}{|c|c|c|c|}
\hline
    &Word length &Word length           &Reduction\\
    & = 21 bits          & = 8 bits     & 8 bits/21 bits\\ \hline
Number of ALUT    &2416     &848     &65\%   \\ \hline
Number of Register  &3114             &576  &82\% \\ \hline
Number of DSP &3              &3  &0\% \\ \hline
\end{tabular}
\end{table}

%\begin{table*}[!t]
%\caption{Performance comparision}
%\label{tab:hw_comparision}
%\centering
%\begin{tabular}{|c|c|c|c|c|c|}
%\hline
%   &\multirow{2}{*}{Proposed hardware} &\multirow{2}{*}{ARM Krait}  &\multirow{2}{*}{Intel Core i5} &Improvement &Improvement\\
%                           &                      &            &                                &compared with ARM  &compared with Intel\\ \hline
%Clock freq.  &\multirow{2}{*}{0.188}   &\multirow{2}{*}{2.26}    &\multirow{2}{*}{2.9} &  \multirow{2}{*}{--}   & \multirow{2}{*}{--}\\
%(GHz)      &     &        &   & &\\ \hline
%Latency  &\multirow{2}{*}{19}   &\multirow{2}{*}{30}    &\multirow{2}{*}{25} &\multirow{2}{*}{37\%} &\multirow{2}{*}{24\%} \\
%(ms)      &     &        &   & &\\ \hline
%Throughput &\multirow{2}{*}{52}  &\multirow{2}{*}{33}   &\multirow{2}{*}{40} &\multirow{2}{*}{58\%} &\multirow{2}{*}{30\%}\\
%(FPS)  & & &  &  &\\ \hline
%Energy  &\multirow{2}{*}{68} &\multirow{2}{*}{75} &\multirow{2}{*}{400} &\multirow{2}{*}{9\%} &\multirow{2}{*}{83\%}\\
%(mJ/frame) &       &       &  & &\\ \hline
%\end{tabular}
%\end{table*}

%\begin{figure}[!h]
%\setlength{\belowcaptionskip}{-0.15in}
%\centering
%\includegraphics[width=3.5in]{ppp.pdf}
%\caption{The relationship between Error and locations of points (Word Length = 10).}
%\label{fig:max_error}
%\end{figure}

Pixels in a patch have various sensitivity to word length truncations. When rotating two different points in a patch by an identical angle, the point that is further from the original point will move further. Hence, the points that is further from the original point will be %more sensitive to truncation errors and be
more likely to be in a wrong location. %Fig. \ref{fig:max_error} shows the relationship between the error defined in equation \ref{eq:error} and locations of original pixels. Results shows that
The maximum errors would occur at points $(\pm15,\pm15)$ which are the farthest points from the original point in a patch.

\begin{figure}[!h]
\setlength{\belowcaptionskip}{-0.00in}
\centering
\includegraphics[width=3.5in]{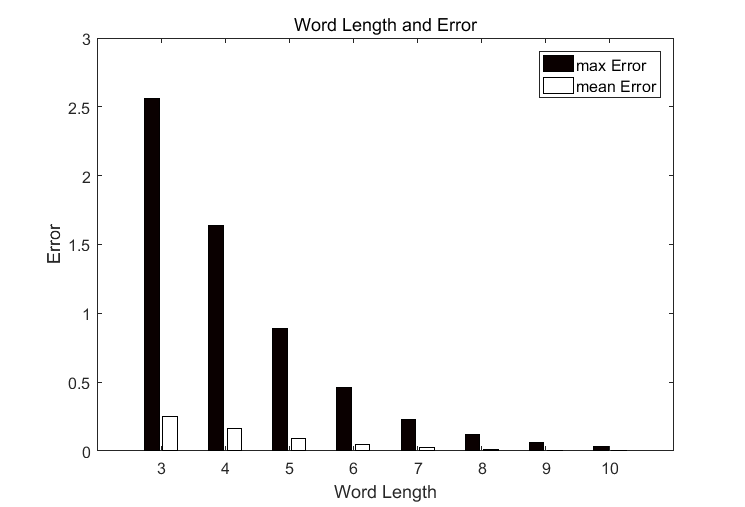}
\caption{The relationship between word length and the maximum Error and mean Error.}
\label{fig:word_len}
\end{figure}

Fig. \ref{fig:word_len} illustrates the relationship between the word length and the maximum and mean errors. The results shows that the maximum error increases exponentially as the word length decreases. Considering both the precision and hardware consumption, the word length of the orientation computing unit is 8 bits in our design.

Table \ref{table:word_len} compares the hardware consumption of the orientation computing module with 21 bits and 8 bits word length. The results show that the numbers of registers and LUTs are significantly reduced by 65\% and 83\% after word length optimization.

\subsection{Hardware evaluation}
The proposed ORB feature extractor is implemented and evaluated on an Altera Stratix V FPGA. %The FPGA device has 71K ALUTs, 6.4MB of BRAM and 352 DSPs.
The proposed hardware consumes 25648 ALUTs, 21791 registers, 1.18M bytes of BRAMs and 8 DSP Blocks. The clock frequency of the proposed hardware is 203MHz. The latency of the hardware for processing an image is 14.8ms, and the througput of the hardware is 67 frames per second.

%To our best knowledge, no work related to hardware ORB feature extractor was pulished before.
We compared the proposed hardware with multi-core ARM processors and Intel CPU. %The ORB algorithm was implemented on ARM Krait platform and Intel Core I5 platform respectively.
Table \ref{tab:hw_comparision} shows the performance comparision between the proposed hardware and ORB implementations on ARM Krait and Intel Core i5 CPU. Compared with ARM Krait, the latency and energy consumption is reduced by 51\% and 9\%, and throughput is improved by 103\%. Compared with Intel i5 CPU, the latency and energy consumption is reduced by 41\% and 83\%, and throughput is improved by 68\%. %Note that the performance of the proposed hardware can be further improved in two directions. For high performance applications, throughput can be improved by more engineering works, such as pipelining and high fan-out nets reduction. Second, for embedded applications, power consumption can be further optimized, such as by optimizing hardware area, clock frequency and supply voltage.

\begin{table}[!t]
\caption{Hardeware performance comparision}
\label{tab:hw_comparision}
\centering
\begin{tabular}{|c|c|c|c|c|}
\hline
                 &Clock Freq.                  &Latency                          &Throughput           &Engergy\\
                 &(GHz)                        &(ms)                            &(FPS)                 &(mJ/frame) \\ \hline
Proposed design     & 0.203                        & 14.8                              & 67             & 68        \\ \hline
ARM Krait           & 2.26                         & 30                              & 33                 & 75        \\ \hline
Intel Core i5       & 2.9                          & 25                              & 40              & 400    \\ \hline
Improvement         & \multirow{2}{*}{--}            & \multirow{2}{*}{51\%}             & \multirow{2}{*}{103\%}  & \multirow{2}{*}{9\%}        \\
vs ARM   &                              &                                 &                       &      \\ \hline
Improvement         & \multirow{2}{*}{--}            & \multirow{2}{*}{41\%}             & \multirow{2}{*}{68\%}  & \multirow{2}{*}{83\%}       \\
vs Intel &                              &                                 &                        &     \\\hline
\end{tabular}
\end{table}

\section{Conclusion}
Feature extraction is the most computation-intensive part in a visual inertial SLAM system.  In our production-level system, we utilize ORB as our feature extractor as ORB is widely used in robotics and it is proven to provide a fast and efficient alternative to SIFT. Based on our profiling results, ORB feature extraction takes over 50\% of the CPU resources as well as energy budget.  In this paper, we aimed to solve this problem and we have designed, implemented, and evaluated a hardware ORB feature extractor. Our design runs %at low frequency,
only at about 203 MHz to reduce energy consumption.  %Even under this low frequency,
In the computation latency category, it outperforms ARM Krait by 51\% and Intel Core i5 by 41\%; in the computation throughput category, it outperforms ARM Krait by 103\%, and Intel Core i5 by 68\%; and most importantly, in the energy consumption category, it outperforms ARM Krait by 10\% and Intel Core i5 by 83\%. Thus this design is proven to be a great balance between performance and energy consumption.

\end{document}